# Sequence to Sequence with Attention for Influenza Prevalence Prediction using Google Trends

Kenjiro Kondo
Aidemy Inc.
University of Tokyo
302 Entrepreneur Plaza, The University of Tokyo, Hongo 7-3-1, Bunkyo-ku, Tokyo
+81-90-1257-3931
kondo-k@aidemy.net

Akihiko Ishikawa
Aidemy Inc.
302 Entrepreneur Plaza, The University of Tokyo, Hongo 7-3-1, Bunkyo-ku, Tokyo
+81-80-1216-6310
ishikawa-a@aidemy.net

Masashi Kimura
Aidemy Inc.
Convergence Lab.
302 Entrepreneur Plaza, The University of Tokyo, Hongo 7-3-1, Bunkyo-ku, Tokyo
+81-90-3095-1760
kimura-m@aidemy.net

## ABSTRACT

Early prediction of the prevalence of influenza reduces its impact. Various studies have been conducted to predict the number of influenza-infected people. However, these studies are not highly accurate especially in the distant future such as over one month. To deal with this problem, we investigate the sequence to sequence (Seq2Seq) with attention model using Google Trends data to assess and predict the number of influenza-infected people over the course of multiple weeks. Google Trends data help to compensate the dark figures including the statistics and improve the prediction accuracy. We demonstrate that the attention mechanism is highly effective to improve prediction accuracy and achieves state-of-the art results, with a Pearson correlation and root-mean-square error of 0.996 and 0.67, respectively. However, the prediction accuracy of the peak of influenza epidemic is not sufficient, and further investigation is needed to overcome this problem.

・`Computing methodologies→Neural networks`

## 1. INTRODUCTION

The prediction of influenza in its early stages reduces its impact along with determining the number of vaccines and other anti-influenza drugs that help the medical personnel to make the correct decision.

Various studies have been conducted to predict the number of influenza-infected people. However, they are not highly accurate, especially in the distant future. To solve this problem, we investigated the sequence to sequence (Seq2Seq) with attention model using Google Trends data to assess and predict influenza prevalence over multiple weeks. The Seq2Seq model allows us to predict multiple time steps prediction. Google Trends data helps to cope with the dark figures, which unreported or undiscovered their statistics. It also provides information about the areas that have been previously searched by users for the prevalence of influenza, which assists in its prediction. In addition, we found that Seq2Seq with attention model achieves state-of-the-art-results.

Our contribution is summarized as follows:

1. We evaluate Seq2Seq models to assess the stability of the prediction of influenza prevalence over multiple weeks.
2. Google Trends data help to compensate the dark numbers including statistics and the improve prediction accuracy.
3. We found that the attention mechanism is significantly effective for the prediction of accuracy. It achieves state-of-the-art performance with a Pearson correlation and root-mean-square error (RMSE) of 0.996 and 0.67, respectively.

The rest of this paper is organized as follows: we summarize the literature related to the prediction of influenza prevalence in section 2. In section 3, we describe the methods used by us. Then, we show the experimental results and discuss them in section 4. Section 6 consists of conclusion remarks.

## 2. RELATED WORKS

Alessa and Faezipour reported a survey of recent studies about influenza detection and prediction. In their survey, they found that most studies and models were developed to detect influenza outbreak from Social Network Service such as the seasonal influenza and swine influenza [1].

Several studies have used the data obtained from Google Flu Trends as the supplemental data. Herman A. C. and Eleftherios M. [2] reported using Google Flu Trends and Google Trends data to predict ILI by statistical analysis. David L., Ryan K., Gary K., and Alessandro V. [3] also reported the prediction of ILI using Google Flu Trends along with the data disclosed by the Centers for Disease Control and Prevention (CDC)[1]. They used the general linear model that achieved a mean absolute error (MAE) of 0.232 during the out-of-sample period. Preis T. and Moat HS. also used the Google Flu Trends. They implemented the mode isolation algorithm and found that when the Google Flu Trends data are combined with historic flu levels, the MAE of in-sample "nowcasts" significantly reduces by 14.4%, in comparison with a baseline model that only uses historic data on flu levels. Ozgur M. A., Dan B., and Robert L. M. [4] used the Google Flu Trends with linear regression models. They found that Google Flu Trends data have a lower RMSE as a predictor variable and the lowest value is achieved when all other variables are included in the model in the forecasting experiments for the first five weeks of 2013 (with RMSE = 57.61). Google Flu Trends data are useful to predict influenza prevalence. However, they are no longer available.

---

[1] https://gis.cdc.gov/grasp/fluview/fluportaldashboard.html

David J. M., and John S. B. [5] used the Wikipedia usage information to predict the prevalence of influenza like diseases in real-time. They developed a Poisson model to predict ILI activity up to two weeks ahead of the CDC, with an absolute average difference between the ILI and the model of just 0.27% for data collected over 294 weeks. They stated that the Wikipedia information accurately estimated the week of peak ILI activity, which was 17% more frequent than Google Flu Trends data and more accurate in its measurement of ILI intensity. However, only the information related to the language of these Wikipedia articles was known, instead of the location of Wikipedia users. José C. S. and Sérgio M used Twitter and web queries. They obtained a precision of 0.78 and an F-measure of 0.83 with the naive Bayes model. However, we used the Google Trends data.

The real-time prediction of ILI prevalence has not been well investigated so far, with the exception of few studies. Hiroshi N. [6] reported the development of a simple method that can be used for real-time epidemic forecasting with a discrete time stochastic model. He stated that real-time forecasting of epidemics has not been widely studied. In this study, a discrete time stochastic model accounting for demographic stochasticity and conditional measurement was developed. This model can derive the uncertainty bounds using chains of conditional offspring distributions. The proposed method was applied to the weekly incidence of pandemic influenza (H1N1-2009) in Japan. Andrea F. D., Mehdi J., Yulia G., Scott L., Fred T., Takeru I., and Richard E. R. [7] used Google Flu Trends with the GARIMA model that predicted weekly influenza cases during seven out-of-sample outbreaks within seven cases for 83% of estimates. The model proposed by us demonstrated a higher prediction accuracy.

Liyuan L. H. and Yiyun Z. W. [8] studied influenza prevalence using long short-term memory (LSTM). They used multiple novel data sources including virologic surveillance, influenza geographic spread, Google Trends, and climate and air pollution to predict influenza trends. However, they did not evaluate the Seq2Seq models.

## 3. METHOD

We used the standard Seq2Seq with attention model to map the variable length input to output. The Seq2Seq model has two main components: encoder and decoder. Usually, encoder and decoder have multiple recurrent units such as LSTM. The attention mechanism determines the time steps that need to be focused on. This method is described in the following subsections.

### 3.1 LSTM

LSTM [9] is a widely used recurrent unit of the Seq2Seq model. Vanilla recurrent neural network [10] are not often efficient in gradient vanishing or explosion problem in the long-time sequence prediction. Additionally, it does not efficiently handle the long-term dependencies. To conquer this problem, LSTM was presented by Hochreiter and Schmidhuber. LSTM comprises a memory cell and three gates: input, output, and forget gates. These gates and the memory cell can record information for a long period, thereby solving the problem of long-term dependencies. The computation flow is described in Figure 1. The output $y^t$ of LSTM is computed as follows:

$$z^t = \sigma(W_z x^t + R_z y^{(t-1)} + b_z)$$
$$i^t = \sigma(W_{in} x^t + R_{in} y^{(t-1)} + b_{in})$$
$$f^t = \sigma(W_{for} x^t + R_{for} y^{(t-1)} + b_{for})$$
$$c^t = i^t \circ +f^t \circ c^{t(t-1)}$$
$$o^t = \sigma(W_{out} x^t + R_{out} y^{(t-1)} + b_{out})$$
$$y^t = o^t \circ h(c^t)$$

Where i, f, and o denote the input, forget, and output gates, respectively, $\sigma(\cdot)$ denotes the sigmoid function, and "∘" denotes the Hadamard product.

The LSTM prediction model is shown in Figure 2. When the feature time sequence input is $X_{1:T}$, LSTM predicts the next time feature $X_{T+1}$. This simple model can predict the feature of the next time step of input.

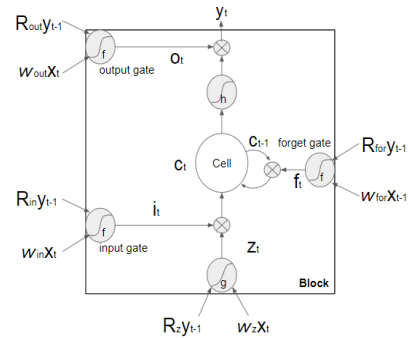

Figure 1. LSTM architecture

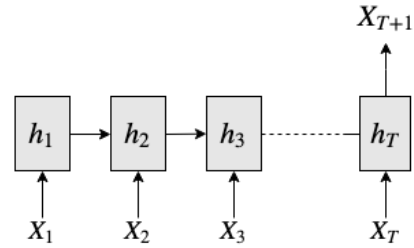

Figure 2. Time series prediction by LSTM

### 3.2 Seq2Seq Model

Unlike LSTM, the Seq2Seq [11] model can predict multiple time steps. This model is illustrated in Figure 3. It consists of two major blocks: encoder and decoder. The encoder outputs the encoded vector of input. The decoder encodes the input vector and predicts the next time step output. Subsequently, if $X_i$ is the input of the next feature sequence, then the LSTM sequence model outputs $X_{i+1}$ as the next time step feature.

Figure 3. Time series prediction by Seq2Seq

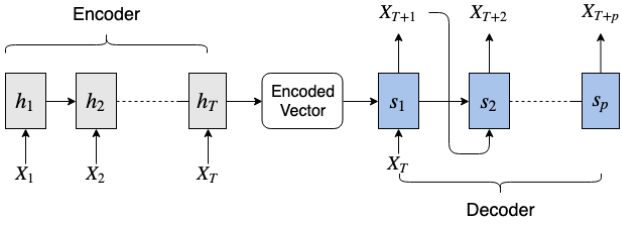

## 3.4 Attention Mechanism

Attention mechanism is a widely used Seq2Seq enhancement, firstly presented in Bahdanau, D., Cho, K., & Bengio, Y [12] to align the word pairs of machine translation source language with the target language. The Seq2Seq with attention model is depicted in Figure 4.

In the Seq2Seq attention model, the predicted feature $X_{T+p}$ is computed as follows:

$$X_{T+p} = g(X_T, h_{enc}, c_i)$$
$$h_{enc} = f(X_{1:T})$$
$$c_i = \sum_{j=1}^{T} \alpha_{ij} h_j$$
$$\alpha_{ij} = \frac{exp(e_{ij})}{\sum_{k=1}^{T} exp(e_{ik})}$$
$$e_{ij} = a(s_{i-1}, h_j).$$

Where, the functions $g(\cdot)$, $f(\cdot)$, and $a(\cdot)$ denote the decoder LSTMs, encoder LSTMs, and densely connected neural network layer with tanh activation, respectively. The probability $\alpha_{ij}$ denotes the attention weights, which reflect the importance of $h_j$ with respect to $s_{i-1}$.

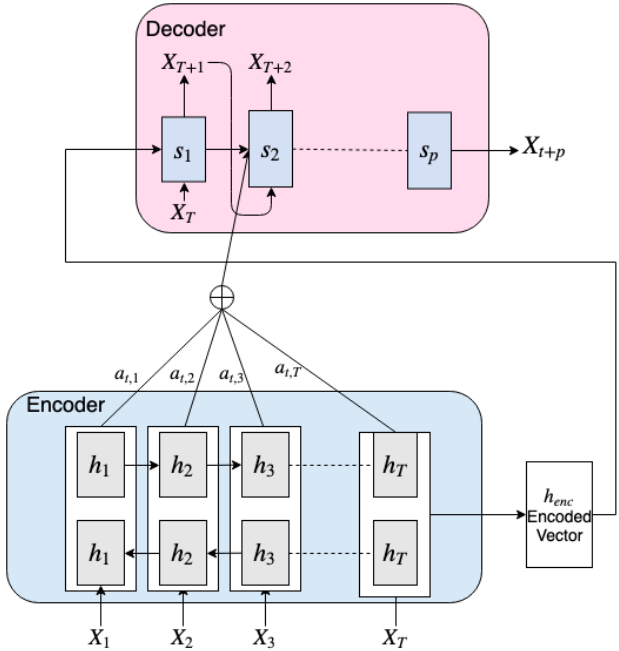

Figure 4. Time series prediction by Seq2Seq with attention

## 3.5 Teacher Forcing

Teacher forcing [13] is frequently used in temporal supervised learning tasks. It replaces the actual output of a unit by the teacher signal in the subsequent computations of the network behavior. It also uses Seq2Seq learning frequently. Let k be a random variable [0, 1] threshold K. Then, the decoder's input signal z(t) is given as

$$z(t) = \begin{cases} X_t & k \geq K \\ Y_t & k < K \end{cases},$$

Where $X_t$ is the actual output of previous LSTM and $Y_t$ is the teacher signal.

## 3.6 Google Trends

Google Trends[2] is a Google service that shows the number of web searches for a specific word. The unweighted ILI collected by the CDC has one-week rags. Moreover, it only represents the cases that have been officially diagnosed by a hospital. In other words, these data have many dark numbers. However, the Google Trends information represents the people's interest at a certain point in time, which can be used to compensate the dark numbers. Therefore, we use the retrieval frequency of the word "influenza" as supplemental information for Seq2Seq with attention. To accomplish this, we concatenated the unweighted ILI and the number of searches for the word "influenza."

## 4. RESULTS AND DISCUSSION

In this section, we illustrate the experimental results of the proposed models and discuss them.

## 4.1 Experimental Conditions

We used the unweighted percentage of the people infected with influenza-like illnesses (unweighted ILI) disclosed by the CDC as the number of people infected by influenza. We collected the unweighted ILI of six states of the US (New York, Oregon, California, Illinois, Texas, Georgia) from the CDC. To assess the prediction accuracy in various climates, we selected these scattered states to prevent similar climate. In addition, we collected the number of times the word "influenza" was searched in these six states from Google Trends. These data were collected between October 10, 2010 and December 30, 2018 (total 430 weeks). We used the first 67% of data as the training data and the remaining 33% as test data. Figure 5 shows how the time series of unweighted ILI was split.

We used the first ten weeks of ILI from the CDC and Google Trends as input and four weeks as output for the prediction of ILI. We performed min-max scaling before the data was given as input. The encoder LSTM was bidirectional with 32 units for all three layers. The decoder LSTM was unidirectional and had a single layer with 64 units. The dropout technique [14] was implemented on both LSTMs to prevent overfitting with a 20% dropout rate. We applied the teacher forcing method with a threshold of 0.8.

## 4.2 Results and Discussion

We evaluated the four methods: SARIMA [15], simple LSTM, Seq2Seq, and Seq2Seq with attention models. The evaluation

---

[2] https://trends.google.co.jp/trends/

metrics considered here were the Pearson correlation coefficient and RMSE.

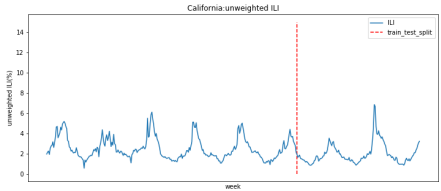

(a) California

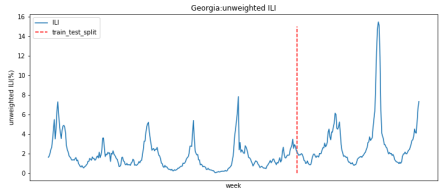

(b) Georgia

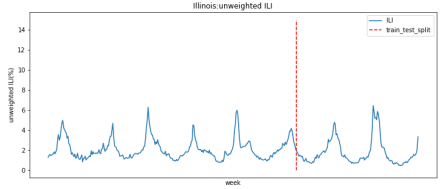

(c) Illinois

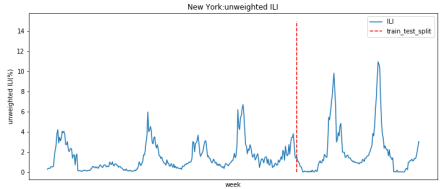

(d) New York

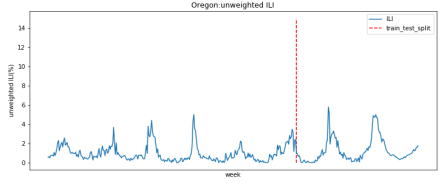

(e) Oregon

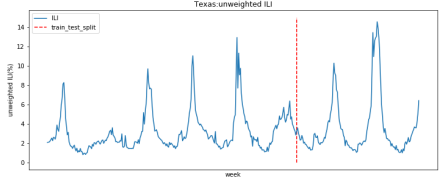

(f) Texas

Figure 5. Unweighted ILI time series split for training and test data for six states. Blue and red lines represent ILI time series and splitting point of training and test data, respectively. Left part of red line is the training data and right part is the test data.

Table 1. Pearson correlation between actual ILI and predicted ILI for each method. Bold words represent the best prediction.

|  | SARIMA | LSTM | Seq2Seq | Seq2Seq with attention |
|---|---|---|---|---|
| California | 0.843 | 0.826 | 0.822 | **0.998** |
| Georgia | 0.869 | 0.884 | 0.900 | **0.991** |
| Illinois | 0.900 | 0.958 | 0.957 | **0.996** |
| New York | 0.880 | 0.904 | 0.894 | **0.997** |
| Oregon | 0.809 | 0.889 | 0.896 | **0.998** |
| Texas | 0.911 | 0.940 | 0.940 | **0.998** |
| Average | 0.869 | 0.900 | 0.902 | **0.996** |

Table 2. RMSE between actual ILI and predicted ILI for each method. Bold words represent the best prediction.

|  | SARIMA | LSTM | Seq2Seq | Seq2Seq with attention |
|---|---|---|---|---|
| California | 0.59 | 0.64 | 0.63 | **0.22** |
| Georgia | 1.40 | 1.42 | 1.37 | **1.20** |
| Illinoi | 0.62 | 0.41 | 0.45 | **0.25** |
| New York | 1.24 | 1.21 | 1.22 | **0.72** |
| Oregon | 0.79 | 0.61 | 0.59 | **0.50** |
| Texas | 1.37 | 1.39 | 1.29 | **0.85** |
| Average | 1.00 | 0.94 | 0.93 | **0.62** |

Table 1 and Figure 6 show the Pearson correlation of each method. The Seq2Seq with attention model shows a significantly higher correlation than other methods for all six states. Table 2 and Figure 7 show the RMSE of each method. Similar to the results of Pearson correlation, the Seq2Seq with attention model shows significantly lower error values than the other three methods.

Figure 8 shows the actual predicted time series of unweighted ILI for one to four weeks. It is evident that the peak value shifts downward as the prediction time increases. The accurate prediction of the peak time is difficult because it cannot be predicted from the learning data when the influenza epidemic subsides. To overcome this problem, further investigation is needed. We consider that the addition of a leading indicator to the input data will enable us to predict the peak of epidemic.

The experimental results show that the attention mechanism is extremely effective to predict the prevalence of influenza. According to our knowledge, a Pearson correlation coefficient of 0.996 is a state-of-the-art result owing to time dependencies. When influenza activity is not at its peak, the long range of ILI depends on prediction. In contrast, during the peak activity season of influenza, the latest values are valid for prediction. The attention mechanism solves this time dependency.

## 5.  CONCLUSION

We demonstrated that the Seq2Seq with attention model is effective in the prediction of influenza prevalence. We achieved state-of-the-art results (0.996 Pearson correlation and 0.67 RMSE) in six states of the US. In addition, the model could predict the prevalence of influenza multiple weeks in advance. However, this prediction is not highly accurate. To overcome this problem, further investigation is needed. We believe that the addition of a leading indicator to the input data will enable us to predict the peak of epidemic.

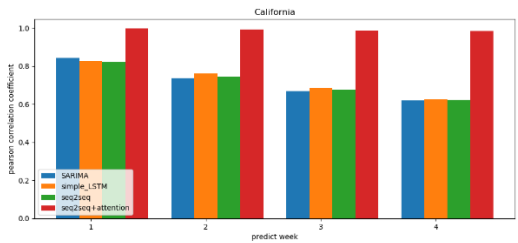

(a)  California

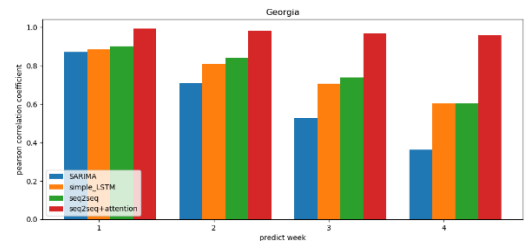

(b)  Georgia

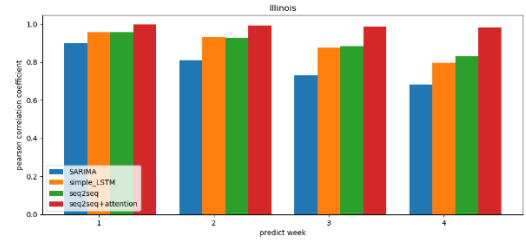

(b)  Illinois

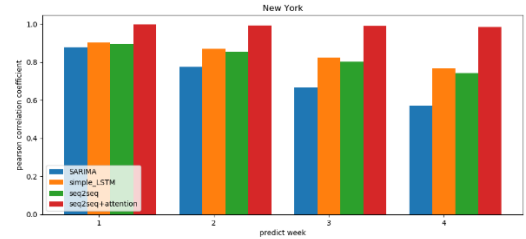

(d) New York

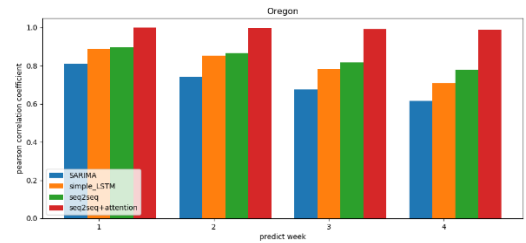

(e)  Oregon

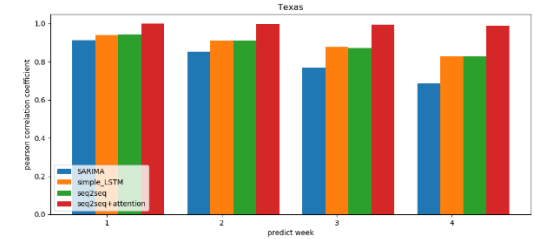

(f)  Texas

Figure 6. Pearson correlation between actual data and predicted data. X-axis represents the predicted week with respect to the input ILI. The blue, orange, green, and red bars represent the base line, simple LSTM, Seq2Seq, and Seq2Seq2 with attention, respectively.

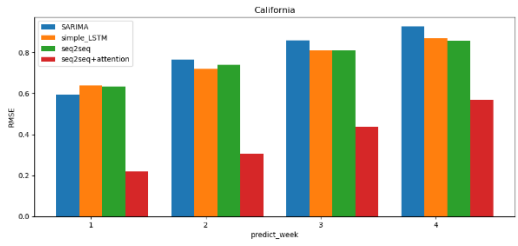

(a) California

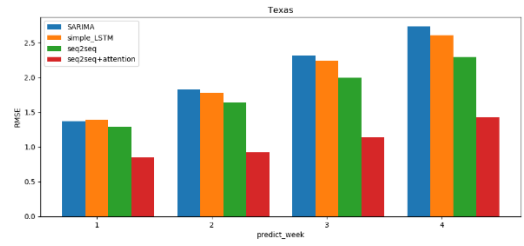

(f) Texas

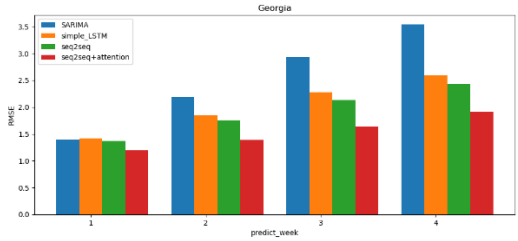

(b) Georgia

Figure 7. RMSE between actual data and predicted data. X-axis represents how the predicted week with respect to the far the predicted week from input ILI. The blue orange, green and red bars represents the base line, simpe LSTM, Seq2Seq, and Seq2Seq with attention, respectively.

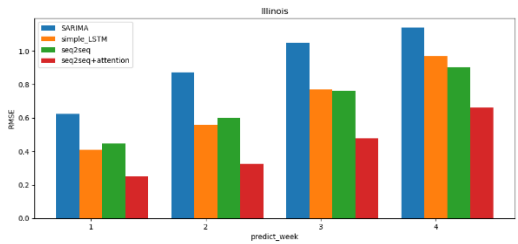

(b) Illinois

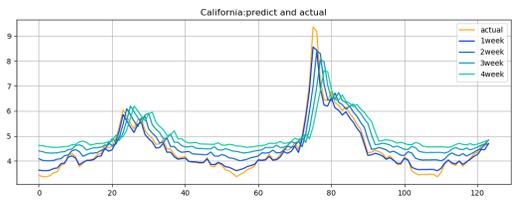

(a) California

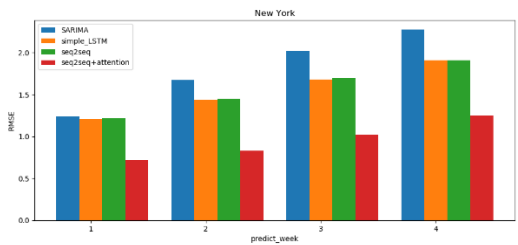

(d) New York

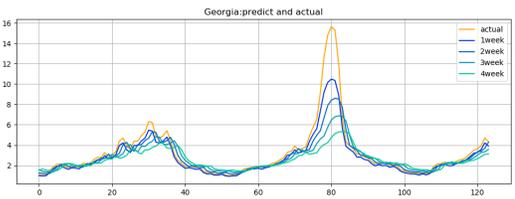

(b) Georgia

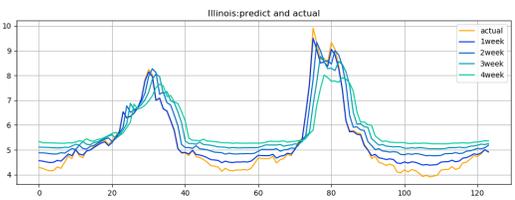

(c) Illinois

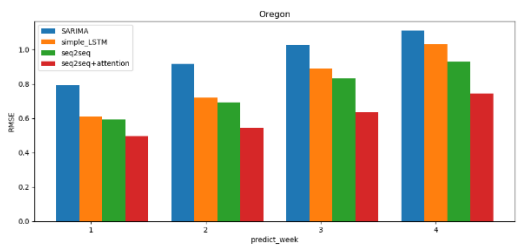

(e) Oregon

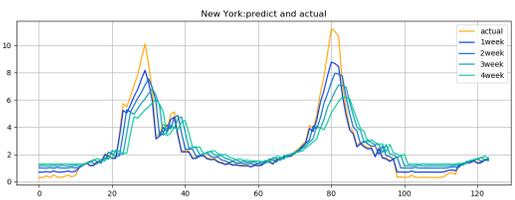

(e) New York

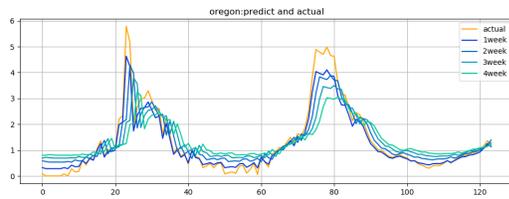

(d) Oregon

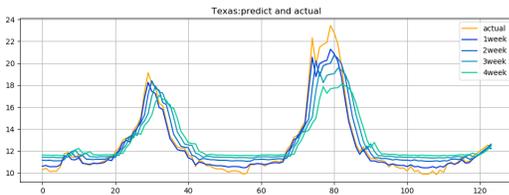

(f) Texas

Figure 8. Predicted time series of unweighted ILI. Orange graph represents actual ILI. Other graphs represent with lag 1 to 4 weeks ahead.